\documentclass[letterpaper, 10 pt, conference]{ieeeconf}  

\IEEEoverridecommandlockouts                              

\usepackage{graphics} 
\usepackage{graphicx}
\usepackage{hyperref}
\usepackage{epsfig} 
\usepackage{amsmath} 
\usepackage{amssymb}  
\usepackage{float}

\title{\LARGE \bf
CLiNet: Joint Detection of Road Network Centerlines in 2D and 3D
}

\author{David Paz$^{*}$, Srinidhi Kalgundi Srinivas$^{*}$, Yunchao Yao$^{*}$, and Henrik I. Christensen$^{*}$
\thanks{$^{*}$Contextual Robotics Institute, University of
    California San Diego, La Jolla, CA 92093, USA {}}%
}

\begin{document}

\maketitle
\thispagestyle{empty}
\pagestyle{empty}

\begin{abstract}

This work introduces a new approach for joint detection of centerlines based on image data by localizing the features jointly in 2D and 3D. In contrast to existing work that focuses on detection of visual cues, we explore feature extraction methods that are directly amenable to the urban driving task. To develop and evaluate our approach, a large urban driving dataset dubbed AV Breadcrumbs is automatically labeled by leveraging vector map representations and projective geometry to annotate over 900,000 images. Our results demonstrate potential for dynamic scene modeling across various urban driving scenarios. Our model achieves an F1 score of 0.684 and an average normalized depth error of 2.083. The code and data annotations are publicly available.\footnote{\href{http://avl.ucsd.edu}{avl.ucsd.edu}} \footnote{\href{https://github.com/AutonomousVehicleLaboratory/clinet}{https://github.com/AutonomousVehicleLaboratory/clinet}}

\end{abstract}

\section{INTRODUCTION}
Open-source contributions to the research community have helped accelerate the development of autonomous vehicle technology in recent years. These contributions include open-source datasets such as Argoverse2 \cite{Argoverse2}, Waymo Open Dataset \cite{sun2020scalability}, and NuPlan \cite{nuplan}. These datasets include labeled sensor and map data that has been curated for various tasks including detection \cite{li2022bevformer, yang2022deepinteraction}, tracking \cite{camo-mot, msmd}, prediction \cite{shi2022mtr, multipath}, and planning \cite{casas2021mp3, mats}. 

A core component of these datasets involves the map information that describes lane level definitions and road network connectivity information that is often utilized in prediction models to provide context and in planning for trajectory generation and optimization tasks. Nevertheless, these models often make a static world assumption that may present failure modes in real-world deployment in the presence of road changes or construction. Examples of methods that utilize high definition maps in autonomy stacks include \cite{autoware, apollo17, micro-mobility} that facilitate global planning for point-to-point navigation; however, given the drastic differences between the definitions, an autonomous agent would be unable to negotiate a feasible trajectory without continuous updates to static map features. This illustrates the importance of dynamic and real-time scene understanding. 

To explore dynamic scene modeling strategies, this work presents three key contributions to the open-source research community, namely,

\begin{itemize}
    \item A publicly available dataset that is automatically labeled by utilizing vector map data from the Argoverse2 sensor dataset. Our work presents an approach for automatic image labeling by leveraging ego-vehicle localization, the state information of each camera, and projective geometry to align map features with image data. The dataset contains over 900,000 annotated image frames across three front facing cameras with various information including centerline and lane boundary definitions. 
    \item A baseline approach termed CLiNet is implemented for centerline key-points prediction with joint 2D and 3D key-point estimation. The approach leverages intrinsic parameters to decouple sensor-specific assumptions from the training process. To the best of our knowledge, this is the first approach that focuses on directly predicting centerline features from camera observations in an urban setting. 
    \item A benchmark is provided to aid in the evaluation of centerline prediction methods. We hope our contribution further motivates the research community to help drive the development of fully dynamic methods for urban driving applications.
\end{itemize}

\begin{figure}
  \begin{minipage}{\linewidth}
    \centering
    \includegraphics[width=\textwidth]{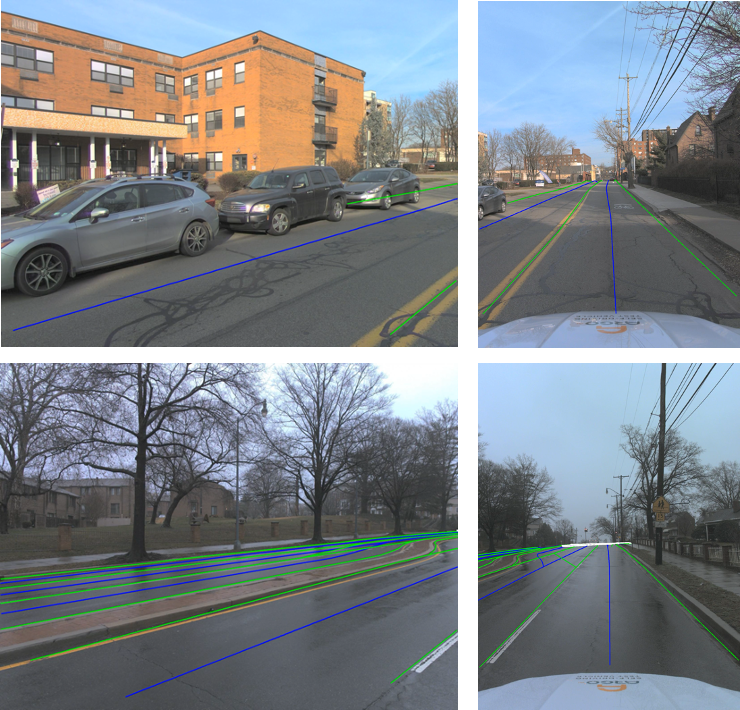}
  \end{minipage}%

  \caption{In contrast to traditional lane detection tasks (shown in green), our work explores an approach for centerline feature prediction, as denoted by blue lines. This work leverages automatic label generation to process a large image dataset from an urban driving scenario with 2D and 3D ground truth information.}
  \label{figure:approach}
\end{figure}

\begin{figure*}
  \centering
  \includegraphics[scale=0.98]{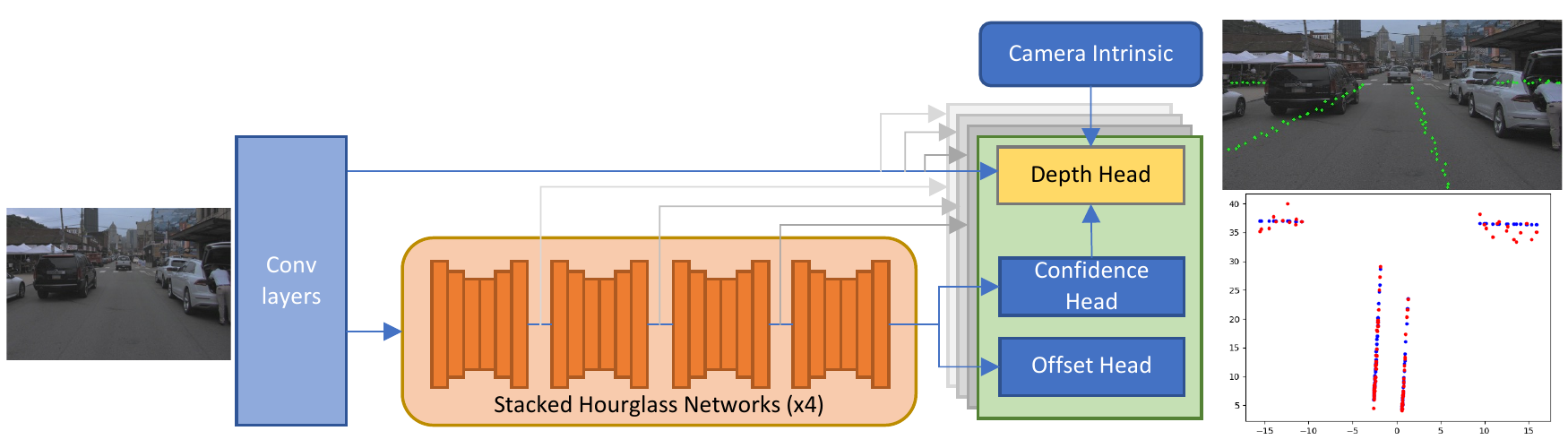}
  \caption{The method introduced utilizes stacked hourglass networks to localize centerline key-points in 2D while generating depth estimates to transform key-point estimates to 3D. Additional details on the hourglass networks can be found in \cite{pinet}}
    \label{fig:architecture}
\end{figure*}

\section{RELATED WORK}
Lane detection and semantic segmentation has remained an active area of research for decades. Early work for autonomous driving applications relied on traditional edge detection techniques, as described in \cite{dickmanns07}. However, with the introduction of learning-based methods for lane detection and segmentation tasks, a number of methods have been developed in recent years. These methods cover semantic segmentation as well as key-point and parametric-based approaches. 

The semantic segmentation task consists of a per-pixel prediction task, in which every pixel from an image is classified into different classes. Public datasets including Mapillary \cite{MVD2017_Mapillary_Vistas} and Cityscapes \cite{Cordts_2016_CVPR_cityscapes} have enabled multiple open-source contributions in the semantic segmentation task that leverage encoder-decoder architectures and spatial pyramid pooling modules \cite{deeplabv3} and attention mechanisms to resolve fine grain details \cite{nvidia-segmentation}. These methods are capable of resolving small details. However, semantic segmentation can present a large computational overhead during training and inference time, which can limit the use cases for real-time scene understanding. 

Another approach for inferring drivable regions and trajectories involves key-point and spline prediction for lanes directly. Examples of real-time image based key-point detection include PINet[15]. In a similar way, the lane detection tasks have been extended to include parameterized spline representations such as Bezier curves \cite{feng2022rethinking} that compress a lane representation by using a fixed number of control points. However, these methods can struggle when recovering irregular lanes and sharp curves that may not be fully characterized by a finite number of control points. More recently, the lane detection task has also been extended to 3D \cite{curveformer} by using synthetic \cite{guo2020gen} and real-world datasets \cite{chen2022persformer}. 

In contrast to existing work, we introduce an approach that focuses on directly predicting centerline key-points as shown in Fig \ref{figure:approach}. A key benefit of directly predicting centerlines involves downstream applications and post-processing. In practice, the centerlines of each lane provide reference trajectories that can be directly utilized by motion planners. Standard methods for detection and segmentation of lanes still infer an overhead for centerline extraction and reasoning depending on the context of the scene and whether a bounded lane mark boundary is traversable or not.



\begin{figure*}
  \centering
  \includegraphics[scale=0.75]{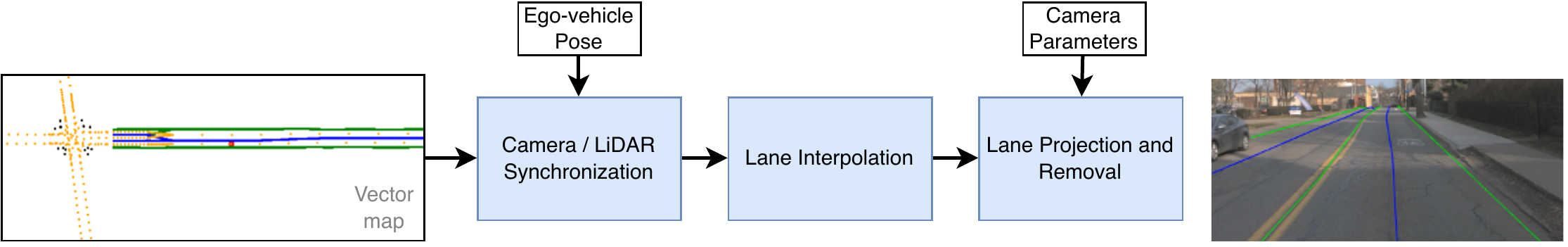}
  \caption{A pipeline is designed to automatically generate image key-point labels that correspond to centerlines via projective geometry. This is accomplished by aligning vector map information (left hand side) regarding each camera on board of the ego-vehicle and projecting 3D features onto an image perspective. Line segments that are far and that correspond to intersection road networks are removed.}
    \label{fig:pipeline}
\end{figure*}

\section{CENTERLINE KEY-POINT PREDICTION}
In this section, we outline the design of CLiNet for the 2D and 3D centerline prediction task. A number of design considerations are made to decouple sensor-specific assumptions from the training process by incorporating the intrinsic parameters of each camera during the depth prediction task. Furthermore, in contrast to semantic segmentation frameworks that seek to classify each pixel with various semantic classes, we simplify the complexity of the model by only predicting the centerline key-points that are directly relevant for the driving task.

\subsection{CLiNet Architecture}
Our approach is motivated by the stacked hourglass network described in the Point Instance Network (PINet) work proposed in \cite{pinet}. This approach stacks multiple hourglass networks during the training process that can be separated during testing as a tradeoff between inference time and precision. The model architecture and training process can be visualized in Fig. \ref{fig:architecture}; where each network jointly learns centerline key-point features and contributes to the overall loss. First, an input image with known intrinsic parameters is normalized and resized to a fixed representation $\mathbf{I_0} \in \mathbb{R}^{H_0\times W_0\times 3}$. A sequence of convolutional layers with batch normalization is then applied to increase channel depth to a final representation $\mathbf{I_1} \in \mathbb{R}^{H_1\times W_1\times C_1}$ such that $H_1 = H_0 / s$, where $s$ is a scaling factor. The prediction heads for each hourglass block further process the intermediate representation $\mathbf{I_1}$ and output a confidence map ($\mathbf{\hat F} \in \mathbb{R}^{H_1\times W_1\times 1}$), depth estimates ($\mathbf{\hat D} \in \mathbb{R}^{N_e \times 1}$), and key-point offset  values ($\mathbf{\hat O} \in \mathbb{R}^{H_1\times W_1\times 2}$) for each key-point on the centerline. For the experiments performed, $C_1=128$, $H_1=256$, $W_1=512$, and $s=8$.  

\subsection{Losses}
The confidence map focuses on the binary classification of key-points, i.e. predictions close to one indicate centerline key-point existence. Given a confidence prediction $\hat F_{i,j}$, the confidence loss is characterized by Eq. \ref{eq:loss}, where the first term minimizes false negatives and the second minimizes false positives. Each term is normalized based on the number of key-points with a ground truth class that corresponds to a key-point $N_e$ and the number of key-points that do not correspond to a key-point $N_d$. A ground truth label is denoted by $F_{i,j}$.

{\small
\begin{equation}
L_{\mathrm {conf }}=\frac{1}{N_e} \sum_{\{i,j | F_{i,j}=1\}}\left(1-\hat F_{i,j}\right)^2 + \frac{1}{N_d} \sum_{\substack{\{i,j | F_{i,j}=0\} }}\left(\hat F_{i,j}\right)^2 
\label{eq:loss}
\end{equation}
}
Together with the confidence map, a two-dimensional offset $\mathbf{\hat O}_{i,j} \in \mathbb{R}^2$ is predicted that indicates a scaling factor to map from a $H_1\times W_1$ representation to the $H_0\times W_0$ input scale. Therefore, the transformed prediction $\hat F_{i,j}$ in original scale, is given by $\left[ j+s\cdot \hat O^x_{i, j}, i+s\cdot \hat O^y_{i,j}\right]^\top$ and $\hat O^x_{i, j}, \hat O^y_{i, j} \in \left [0, 1\right ]$. The offset loss is then characterized by

\begin{equation*}
\begin{aligned}
L_{\mathrm{offset}}&=&\frac{1}{N_e} \sum_{\left\{i, j \mid F_{i, j}=1\right\}}\left(O^x_{i,j}-\hat{O}_{i, j}^x\right)^2 + \\
&&\frac{1}{N_e} \sum_{\left\{i, j \mid F_{i, j}=1\right\}}\left(O^y_{i,j}-\hat{O}_{i, j}^y\right)^2 
\end{aligned}
\end{equation*}

Given that the features are encoded in 2D, an additional multilayer perceptron (MLP) is utilized to infer depth for each centerline key-point. A depth prediction $\hat Z_{i,j}$ is estimated by first converting a key-point prediction in pixel space $\left [i,j \right] ^\top$ to camera coordinates. This transformation is given by the intrinsic camera parameters and by accounting for image resizing and crop operations. The camera-centric coordinates $\left [X_{i.j}, Y_{i,j} \right] ^\top$ are then concatenated with its corresponding image embedding $\mathbf{I_1}_{i,j}$ to provide additional context. This final representation is passed through the MLP to predict its z-component, namely $\hat Z_{i,j}$. The depth loss is then characterized as shown below, where $Z_{i,j}$ denotes ground truth.

\begin{equation*}
L_{\mathrm {depth}}=\frac{1}{N_e} \sum_{\left\{i, j \mid F_{i, j}=1\right\}}\left(Z_{i, j}-\hat{Z}_{i, j}\right)^2 
\end{equation*}

In summary, the losses are added and optimized jointly during the training process using the Adam optimizer with a constant learning rate. The overall loss is given by $L=L_{\mathrm {conf }} + L_{\mathrm{offset}} + \gamma L_{\mathrm {depth }}$, where $\gamma$ is a constant.

\section{EXPERIMENTS}
To evaluate our reference solution, we introduce a large dataset for training, validation, and testing based on the Argoverse2 sensor data~\cite{Argoverse2}. The evaluation additionally consists of a benchmark to quantify the performance of the approach for key-point localization in 2D and depth estimation (3D). Additional details on the data and benchmarks are provided below.

\subsection{Dataset}
We augment the dataset to create a new reference termed AV Breadcrumbs that consists of $942,161$ individually labeled image frames across three front-facing cameras with known camera parameters. In contrast to standard labeling techniques that generally involve extensive manual labeling, we design a pipeline that automatically generates image key-point features with ground truth depth information. First, the ego-vehicle 
pose is synchronized to each camera by utilizing the LiDAR; given the pose of each camera over time and performing pose corrections, we align vector map information defined in a city coordinate frame and project its 3D features into each image frame. Given that projection preserves 2D and 3D correspondences, this facilitates the process of assigning unique lane IDs. 

\begin{figure*}
  \centering
  \includegraphics[scale=0.16]{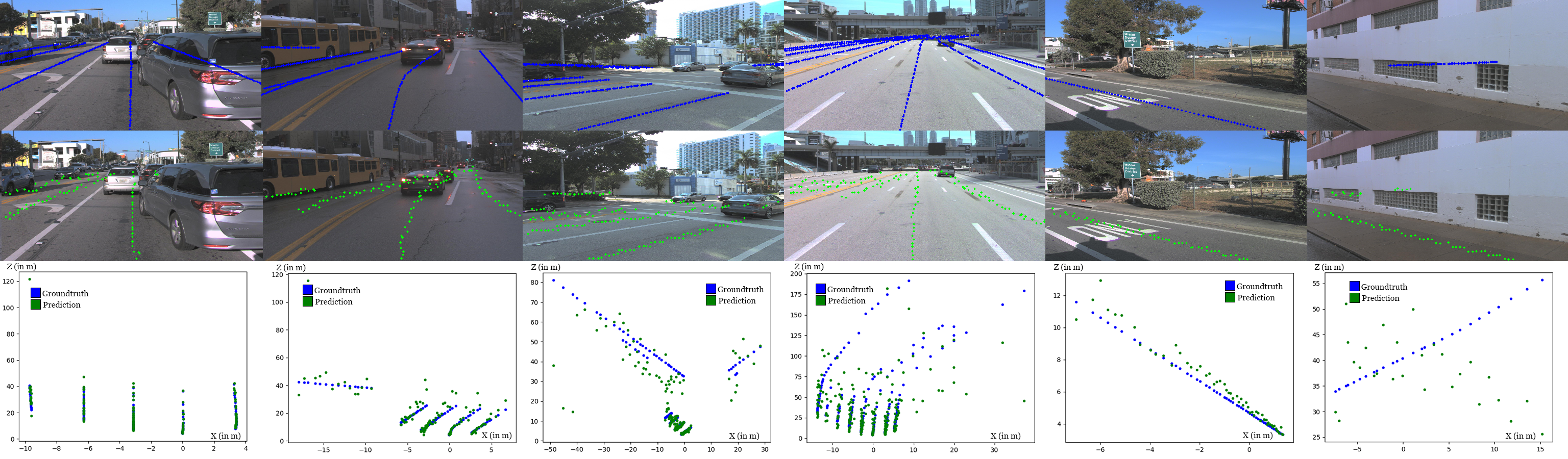}
  \caption{Data visualizations correspond to ground truth 2D labels (top row), 2D predictions from our approach (CLiNet - middle row), as well as depth estimates where the vertical axis corresponds to the z-axis and the horizontal axis corresponds to the x-axis regarding each camera frame~(bottom row). Predictions and ground truth labels are denoted in green and blue, respectively.}
\label{fig:data}
\end{figure*}

The dataset includes labels for centerlines and lane boundaries; where each label is interpolated in 3D and post-processed in an image frame to remove features that are too far or too tightly packed after projection. An important consideration involves intersections: although the full road network connectivity for each intersection is available, we find that sufficient context to identify entry and exit points of each intersection lane is often lacking when the vehicle is at an intersection. For this reason, we additionally remove labels that are too short or are part of intersections. The total split for the train, validation, and test sets corresponds to 659,720, 141,138, and 141,303 image frames, respectively. The auto-labeling pipeline is shown in Fig. \ref{fig:pipeline} and a few data labels can be observed in the first row of Fig. \ref{fig:data}. On average, we observe adequate alignment between the automatically generated features and the lanes visualized in each image.

\subsection{Results}

Each of the logs in the training, validation and test set comprises of sequential frames of data. We take a subset of these logs for our training and testing as two consecutive frames of data present similar features. As a result of which, we empirically chose every 10th image in the sequence for training and testing. This reduced the number of test images to 13,046 images.

Fig: \ref{fig:data} shows the results from our model for the test set of our dataset. It can be seen that our model performs well in detecting the key-points on the straight line segments (row:2, col:4) as well as curved line segments (row:2, col: 2). Additionally,  we observe that there are some challenging scenarios due to lack of context (col:1)  and complexities due to complete occlusions (col:6).
 
The third row of the figure illustrates predicted depth versus ground truth depth for each of the input images in the top row. We observe that the points that are at a greater distance from the camera center have relatively large error compared to those that are closer to the camera center (row:3, col:4).

Table \ref{tab:benchmark} shows the average precision, recall and F1 score of our model on the test set. We create an occupancy grid map of the predicted key-points where the centerline key-point indices are set to 1. To calculate the number of true positives, false positives and false negatives, we then create a window of size \textit{n} x \textit{n}. If the occupancy grid does not contain a key-point within the windowed region, we count this key-point to be a false negative. We zero out the windowed region otherwise, which we later use for counting false positives.

To calculate the average depth error, we use the L1 norm between the ground truth and the predicted \textbf{z} values. We normalize this error based on the distance of the key-point from the camera center. Error for key-points closer to the camera center is penalized more than the error for key-points farther away. Average normalized depth error for the test set is \textbf{2.0834}.

\begin{table}[htbp]
\caption{Benchmark Results on the Test Set}
\begin{center}
\begin{tabular}{|c|c|c|c|}
\hline
\textbf{Window Size} & \textbf{Precision} & \textbf{Recall} & \textbf{F1 Score} \\
\hline\hline
5 & 0.822 & 0.585 & 0.684  \\
\hline
3 & 0.748 & 0.512 & 0.608  \\
\hline
1 & 0.378 & 0.229 & 0.285  \\
\hline
\end{tabular}
\label{tab:benchmark}
\end{center}
\end{table}


\section{CONCLUSION AND FUTURE WORK}
In this work, we introduced a large-scale dataset for 2D and 3D centerline detection. A reference solution to help drive progress for future contributions was additionally presented as a baseline that shows potential in real-time dynamic scene understanding. 

We plan to explore further methods that improve depth perception and enforce temporal consistency. We also plan to perform curve fitting through the 2D key-points and provide additional metrics for accuracy and error. There are cases where the lane information cannot be extracted from the image due to occlusions that are a result of reprojection as show in the last column of Fig. \ref{fig:data}. We plan to handle these cases for future work by leveraging depth data from sensors. Additionally, given the practicality of our labeling pipeline, we plan to release additional data to augment the training process and further generalize for other scenarios and sensor configurations.






\section*{ACKNOWLEDGEMENT}
We appreciate the input from various members of the Autonomous Vehicle Lab, specifically Henry Zhang, Rohin Garg and Tanay Karve for their feedback on open-source driving datasets. Our team also thanks the Argo AI team for their contributions to the Argoverse2 open-source dataset. This work would not have been possible without their contributions.



\bibliographystyle{./IEEEtran}
\bibliography{IEEEcitation}

\end{document}